\documentclass[hidelinks,onefignum,onetabnum]{siamart220329}

\usepackage[english]{babel}
\usepackage{float}
\usepackage{amssymb,amsmath}
\usepackage{array,booktabs,arydshln,xcolor}
\usepackage{bbm}
\usepackage{graphicx}

\newcommand\VRule[1][\arrayrulewidth]{\vrule width#1}
\newcommand{\norm}[1]{\left\lVert#1\right\rVert}
\newcommand{\probP}{\text{I\kern-0.15em P}}
\renewcommand{\d}{\mathrm{d}}
\renewcommand{\epsilon}{\varepsilon}

\usepackage{amsfonts}
\usepackage{graphicx}
\usepackage{epstopdf}
\usepackage{algorithmic}
\ifpdf
  \DeclareGraphicsExtensions{.eps,.pdf,.png,.jpg}
\else
  \DeclareGraphicsExtensions{.eps}
\fi

\newsiamremark{remark}{Remark}
\newsiamremark{hypothesis}{Hypothesis}
\crefname{hypothesis}{Hypothesis}{Hypotheses}
\newsiamthm{claim}{Claim}

\headers{GANs and Closures}{E. R. Crabtree, J. M. Bello-Rivas, A. L. Ferguson, and I. G. Kevrekidis}

\title{GANs and Closures: Micro-Macro Consistency \\
in Multiscale Modeling\thanks{Received by the editors August 24, 2022; accepted for publication (in revised form) April 18, 2023; published electronically DATE.
\funding{This work was partially supported by Multi-University Research Initiative grants by the US Army Research Office and the by the
US Air Force Office of Scientific Research.}}}

\author{Ellis R. Crabtree\thanks{Department of Chemical and Biomolecular Engineering, Johns Hopkins University, Baltimore, MD 21218 USA 
  (\email{ecrabtr2@jhu.edu.com}, \email{jbellor2@jhu.edu, yannisk@jhu.edu}).}
\and Juan M. Bello-Rivas\footnotemark[2]
\and Andrew L. Ferguson\thanks{Pritzker School of Molecular Engineering, University of Chicago, Chicago, IL 60637 USA
  (\email{andrewferguson@uchicago.edu}).}
\and Ioannis G. Kevrekidis\footnotemark[2]}

\usepackage{amsopn}
\DeclareMathOperator{\diag}{diag}

\begin{document}
\maketitle

\begin{abstract}
Sampling the phase space of molecular systems --and, more generally, of complex systems effectively modeled by stochastic differential equations (SDEs)-- is a crucial modeling step in many fields, from protein folding to materials discovery. These problems are often multiscale in nature: they can be described in terms of low-dimensional effective free energy surfaces parametrized by a small number of ``slow'' reaction coordinates; the remaining ``fast'' degrees of freedom populate an equilibrium measure {\em conditioned} on the reaction coordinate values. Sampling procedures for such problems are used to estimate effective free energy differences as well as ensemble averages with respect to the conditional equilibrium distributions; these latter averages lead to closures for effective reduced dynamic models. Over the years, enhanced sampling techniques coupled with molecular simulation have been developed; they often use knowledge of the system order parameters in order to sample the corresponding conditional equilibrium distributions, and estimate ensemble averages of observables. An intriguing analogy arises with the field of machine learning (ML), where generative adversarial networks (GANs) can produce high-dimensional samples from low-dimensional probability distributions. This sample generation is what is called, in equation-free multiscale modeling, a ``lifting process'': it returns plausible (or realistic) high-dimensional space realizations of a model state, from information about its low-dimensional representation. In this work, we elaborate on this analogy, and we present an approach that couples {\em physics-based} simulations and biasing methods for sampling conditional distributions with {\em ML-based} conditional generative adversarial networks (cGANs) for the same task. The ``coarse descriptors'' on which we condition the fine scale realizations can either be known {\em a priori} or learned through nonlinear dimensionality reduction (here, using diffusion maps). We suggest that this may bring out the best features of both approaches: we demonstrate that a framework that couples cGANs with physics-based enhanced sampling techniques can improve multiscale SDE dynamical systems sampling, and even shows promise for systems of increasing complexity (here, simple molecules). 
\end{abstract}



\section{Introduction}
\label{sec:intro}
An important goal of the study of (multiscale) dynamical systems is the exploration and understanding of the system behavior over time scales
of interest --typically, of the long term system dynamics. 
Fine scale dynamical system models, with many degrees of freedom, often must be (for practical prediction purposes) 
reduced to significantly lower-dimensional representations that describe
the features of the behavior relevant to the macroscopic scale observer/modeler. 
Identifying and using such a reduced model, starting from models at the 
microscopic level, is a truly challenging task. 
Finding the ``right observables/variables'' for these representations has
been a holy grail of multiscale modeling and data analysis for more than 100
years - from principal component analysis to autoencoders (or ``nonlinear principal components'' when they were first proposed) to isomap, local linear embedding, diffusion maps, and a host of nonlinear data mining/manifold learning techniques \cite{pca, autoencoder, isomap, locallinear, COIFMANdmaps}.

Beyond such computational, data-driven techniques that 
intelligently reduce the system to a low-dimensional representation, there
is also a clear need for the converse: techniques that ``lift,'' or in other words, techniques that return the system from
coarse representations to consistent, and plausible, fine representations, thus returning the system state conditioned on the reduced observables back to the high-dimensional space.
It is immediately obvious that this inverse map is multivalued: there are countless conformations of molecules of gas confined to a box that have the same pressure and temperature --some of them more probable than others.
It is also important to note that there is a vast variability in the grades of computational difficulty across such problems: it is easy to lift from pressure and temperature to gas
molecule positions and velocities in a box, but it is much more challenging to construct a molecular configuration with a prescribed pair-correlation function.

Mathematical/computational methods that accomplish this conditional ``lifting''
have existed --and are used daily-- in  traditional first principles modeling,
from initializing on a slow manifold 
of a kinetic network
to umbrella sampling (US) 
for molecular dynamics \cite{TORRIE1977}.
Yet recent developments in the field of machine learning have opened the 
door to new possibilities. 
Generative adversarial networks (GANs) \cite{goodfellow2014} and other recent generative models that are fundamentally different from GANs \cite{diffmodels1, diffmodels2, diffmodels3, SOIZE2016242, plom} have demonstrated considerable successes in 
generating high-dimensional data from low-dimensional probability distributions.
For instance,
a GAN architecture is capable of creating images that appear like real photographs of human faces; 
a GAN creates photos with unique, detailed, and realistic subtle texture/features that are consistent with 
the coarse features that make up a realistic human face \cite{karras2019stylebased}. 
Furthermore, an extension of the original GAN architecture --the conditional generative adversarial network (cGAN) \cite{mirza2014} -- can sample from 
conditional distributions to produce high-dimensional data conditioned on given training values, which provides further information (in the example of human faces, this would amount to having a 
GAN that can sample based on prescribed criteria such as eye color, hair color and length, ear size, etc.). 

\begin{figure}[ht]
    \centering
    \includegraphics[width=1\linewidth]{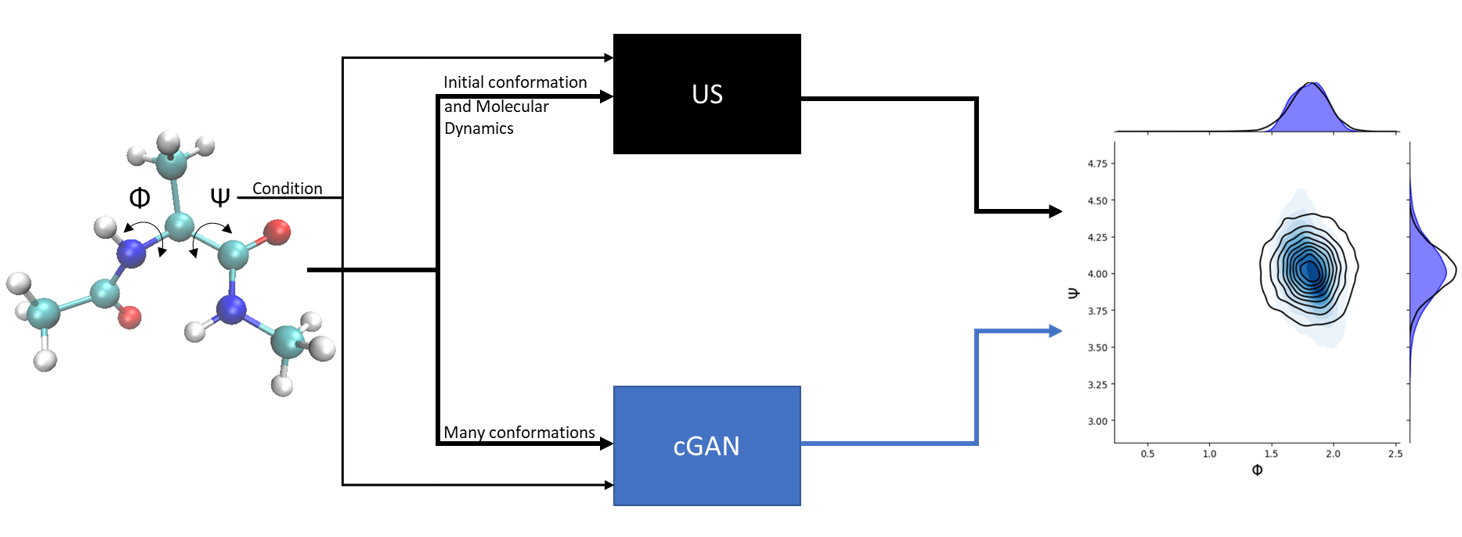}
    \caption{An illustrative schematic showing that, in the context of the molecular simulation of alanine dipeptide, a cGAN conditioned at a particular value of a collective variable (in this case, the dihedral angle represented by $\Psi$) produces results not unlike an Umbrella Sampling (US) simulation biased at the same value of $\Psi$. The data generated by a cGAN (blue) is colored in shades of blue by empirical density, while the empirical density of the US produced data is represented by black contour lines. The empirical marginals of the US and cGAN outputs with respect to the collective variables of the alanine dipeptide are shown on the margins of the plot.}
    \label{fig:pr_fig}
\end{figure}

While GANs are certainly an exciting development, they exhibit parallels with well-understood techniques from the fields of statistical mechanics and dynamical systems. 
In the area 
of molecular simulation, techniques such as free energy perturbation, umbrella sampling, adaptive 
biasing force, metadynamics, and newer methods such as intrinsic map dynamics can respectively be used to (a) calculate 
free energy differences between coarse states of a system 
and, subsequently, (b) bias further molecular simulations in order to fully sample configurational space 
along a set of variables that describe the bulk behavior of a system --what are often called collective variables (CVs) \cite{TORRIE1977,FEPZwanzig,Adaptbiasforce,Metad,imapd}. 
At their core, these biased simulations 
generate high dimensional molecular data conditioned on CVs of choice. 
Figure \ref{fig:pr_fig} displays a general schematic showing that both cGANs and enhanced sampling techniques such as US accomplish the same task of producing data that is consistent with some prescribed CV.

The goal of this work is 
to improve the understanding of the similarities between GANs and other established biased sampling techniques in 
physics and chemistry and, importantly, to develop a framework for \emph{utilizing them in tandem}. 
On the one hand, physics-based simulation and sampling methods often suffer from slow convergence times, whereas GANs can quickly and efficiently sample from a target distribution once trained, sidestepping 
the problem of slow convergence. 
On the other hand, some drawbacks of GANs are the (typical) absence of physical constraints or priors within the generative ``lifting'' process and the requirement
{\em to be trained offline on large data sets}. 
We believe (and suggest, and illustrate how) the second of these problem could be mitigated by coupling GANs with 
physics-based sampling methods that generate training data adaptively from the original, high-dimensional 
dynamical system on demand. 
In a sense, the available fine scale physics help construct on the fly --and adaptively-- the large training set that enables GAN efficiency. 

These methods, new and old, complement each other 
and have the potential to formulate a more efficient sampling workflow when used together. In this 
workflow, GANs learn low-dimensional probability distributions and lift to high-dimensional 
space, expediting the exploration of the phase space of the multiscale dynamical system. 
Neural network based generative models have been shown to be successful in identifying CVs \cite{stoltz2022} and also interpolating simulation frames between molecular conformations \cite{mlss}. The framework proposed in this work not only identifies the slow directions (or CVs) of the multiscale dynamical system in a data-driven manner, but also produces the conditional equilibrium measure of the system at a desired value of the CV(s). Furthermore, this framework is not limited to solely dynamical systems described by SDEs, as GANs can be applied to any problem that requires, or benefits from, lifting. \par

The paper is organized as follows: Section \ref{sec:framework} proposes the general framework of the coupling method in algorithm format. Section \ref{sec:US_on_dmap_CVS} demonstrates physics-based conditional sampling on known variables and physics-based sampling directly on learned variables from dimensionality reduction. 
Section \ref{sec:cGAN_sampling} describes conditional generative adversarial networks (cGANs) as a machine learning based alternative to physics-based sampling. 
Section \ref{sec:gan-assisted} illustrates how coupling cGANs with physics-based sampling can expedite sampling of a dynamical system's phase space. 
Section \ref{sec:alanine} compares results of cGANs applied to molecular dynamics simulation data with physics-based methods. 
Section \ref{sec:dim_and_extrap} demonstrates additional uses and capabilities of cGANs when applied to multiscale dynamical systems. It is in this context
that the use of cGANs towards effective SDE closures arises.
This gave rise to the pun in our paper's title, reminiscent of the name
of an iconic American hard rock band. 
Section \ref{sec:conclusions} summarizes the results (and possible shortcomings) of the work and outlines possibilities for further applications.
\section{Proposed framework}
\label{sec:framework}
Consider a stochastic differential equation (SDE) with equilibrium distribution $P$.
In other words, in the long time limit, points drawn from a trajectory $\{ Z_t \mid t \ge 0 \}$ of the SDE are distributed according to $P$.

Applying manifold learning to the points of a sufficiently long trajectory $Z_t$, we can identify a set of slow variables $Y$ of the SDE so that a point $Z$ drawn from $P$ can be written as $Z = (X, Y)$ in some system of coordinates.

We propose a method for sampling the conditional measure $P(X \mid Y = y)$ that consists of the following:
\begin{enumerate}
\item Training a conditional GAN, $\hat{P}(X \mid Y = y)$, so that it approximates $P(X \mid Y = y)$ for arbitrary $y$.
  Training is done only once and its computational expense is meant to be amortized.
\item Generating $N$ initial points along $\{ (X, Y) \mid Y = y \}$ using the cGAN.
\item Propagating the initial points using a modified SDE (umbrella sampling) that enforces that the trajectories lie in $\{ (X, Y) \mid Y = y \}$.
  This step can be trivially parallelized, as each trajectory is independent of the others.
\item Combining the points of all of the $N$ conditional trajectories using a histogram reweighting method.
\end{enumerate}
In Section~\ref{sec:gan-assisted}, we exhibit numerical evidence that indicates that as long as the conditional distributions generated by the cGAN are not too far away from the true conditionals, then our proposed approach is faster and less computationally expensive than the alternative approach consisting of sampling $\{ X \mid Y = y \}$ by a single long run of a (modified) SDE.

\section{Physics-based conditional sampling on known collective variables and on learned ones}
\label{sec:US_on_dmap_CVS}
In order to ``lift'' from coarse scale variables (variables that parametrize the slow, lower-dimensional manifold on which the coarse system dynamics evolve) to realistic, consistent, fine scale realizations/variables, one must either know the coarse variables {\em a priori} or learn them from data through dimensionality reduction. As an illustration, consider the dynamical system described by the two-dimensional system of SDEs below:
\begin{equation}
    \begin{split}
    &dx_1(t) = a_1dt + a_2dB_1(t),\\
    &dx_2(t) = a_3(1-x_2)dt + a_4dB_2(t),
    \end{split}
    \label{eq:untransformed_SDE}
\end{equation}
where $B_1$ and $B_2$ are standard Brownian motions, $a_1 = a_2 = 10^{-3}$, and $a_3 = a_4 = \epsilon{a_1} = \epsilon{a_2} = 10^{-1}$, where the ratio of time scales is $\epsilon = 10^2$;
these scalings imply that $x_1$ is the slow variable and $x_2$ is the fast variable ($x_2$ is faster than $x_1$ by a factor of $10^2$). Information on how this system of SDEs (as well as the other dynamical systems described by SDEs in this work) was integrated can be found in Appendix \ref{sec:app-d}. To facilitate exploration of the phase space, physics-based methods can be employed to bias the system along the slow variable, $x_1$.
Umbrella sampling (US) is a popular method in which series of biased simulations are used to overcome free energy barriers in the computation of effective free energy surfaces in molecular dynamics \cite{TORRIE1977}. 
US involves the superposition of a harmonic potential, $U(x_1, x_2) = \frac{k}{2}(x_1-x_{1_0})^2$ with the original system potential, for the purpose of steering the biased simulations towards certain prescribed values of the reaction coordinates of the system. Here the variable in this harmonic biasing potential
is the full system state, $(x_1,x_2)$, $k$ is a spring constant governing the intensity of the bias, $x_1$ is the reaction coordinate around which the bias will be applied, and $x_{1_0}$ is the prescribed value of this variable. In this example, if $x_1$ is known to be the slow variable, the bias can be added directly in the form of the force: $F(x_1,x_2) = -\nabla U(x_1,x_2) = (-k(x_1-x_{1_0}),0)$, resulting in a new system of equations:
\begin{equation}
    \begin{split}
    &dx_1(t) = (a_1-k(x_1(t)-x_{1_0}))dt + a_2dB_1(t),\\
    &dx_2(t) = a_3(1-x_2(t))dt + a_4dB_2(t).
    \end{split}
\end{equation}
Regarding the force constant $k$, the strength or magnitude of $k$ will determine the range of the slow variable explored in the simulation. A weaker force constant will allow exploration of a larger span of slow variables. This choice is discretionary and is dependent on what is wanted or required of the user. Additionally, the bias added to the system is not required to be quadratic, and other biases (for example, a linear bias) can be added depending on the slow variable behavior desired in the biased simulation. A quadratic bias on the slow variable is generally used to force the slow variable to the desired coordinate in both directions, restricting it to values within the parabola described by the bias. It is also worthwhile to note that this added bias then alters the conditional distribution of the slow variable. If the true distribution of the slow variable is of interest, reweighting techniques such as the weighted histogram analysis method (WHAM) \cite{wham} and the multistate Bennett acceptance ratio (MBAR) \cite{mbar} should be used to recover the true distribution of the slow variable.

If the slow variable is not known, a slow coordinate can be identified through data-driven dimensionality reduction techniques. In our case, the nonlinear dimensionality reduction technique of choice is diffusion maps \cite{COIFMANdmaps}, the use of which leads to data-driven coordinates that parametrize the slow manifold of the system dynamics.
Diffusion maps accomplish this by using a diffusion process to explore the underlying geometry of the data and subsequently identify high-variance variables (or variable combinations) that parametrize a low-dimensional manifold of the data. Further elaboration on diffusion maps can be found in Appendix \ref{sec:app-c}.
%
In our illustration, a single such diffusion map coordinate is used to construct the biasing potential: $U(\textbf{X}) = \frac{k}{2} \left( \phi(\textbf{X}) - \phi_0 \right)^2$, where $\textbf{X} = (x_1,x_2) \in \mathbb{R}^2$ is the SDE full state, $\phi = \phi(\textbf{X})$ is the first diffusion map coordinate parametrizing the slow direction, and $\phi_0 \in \mathbb{R}$ is the constant value towards which the system will be conditionally biased. 
In general, $\phi = \phi(\textbf{X})$ is not an explicit function of the system state (like $x_1 = x_1(\textbf{X})$ here); it is a vector $\boldsymbol{\phi} \in \mathbb{R}^N$, where $N$ is the number of data samples. One needs to use appropriate interpolation techniques, like geometric harmonics \cite{COIFMANgeoharmonics, GPDMAPFelix} or Gaussian process regression \cite{GProcessesRasmussen} to
express a fitted function, $\Phi_1(\textbf{X})$, to interpolate the data $\{ (\textbf{X}_i, \boldsymbol{\phi}_i) \in \mathbb{R}^n \times \mathbb{R} \mid i = 1, \dotsc, N \}$, where $n$ is the number of variables and $N$ is the number of samples.
Once $\Phi_1(\textbf{X})$ is obtained, it can be used in lieu of $\phi$ in the biasing potential. Additionally, gradients of the function $\Phi_1(\textbf{X})$ with respect to $\textbf{X}$ (here $x_1$ and $x_2$) can be computed through direct calculation of the derivatives or automatic differentiation, allowing us to implement the bias directly on the diffusion map coordinate. (We observe that other dimensionality reduction schemes are, of course, also viable \cite{Wang_Ferguson_dimred}. Autoencoding neural networks are a particularly attractive alternative due to the intrinsic provision of gradient information and interpolative capacity \cite{VAE}.)
Thus,
$$U(\textbf{X}) = \frac{k}{2} \left( \Phi_1(\textbf{X}) - \phi_0 \right)^2$$
and the corresponding force is
$$F(\textbf{X}) = -\nabla U(\textbf{X}) = -k \nabla\Phi_1(\textbf{X})(\Phi_1(\textbf{X}) - \phi_0)$$
resulting in a new, final set of biased SDEs for our example:
\begin{equation}
    \begin{split}
    &dx_1(t) = (a_1-k(\Phi_1((x_1(t),x_2(t))-\phi_0)\frac{\partial \Phi_1}{\partial x_1})dt + a_2dB_1(t),\\
    &dx_2(t) = (a_3(1-x_2(t))-k(\Phi_1((x_1(t),x_2(t))-\phi_0)\frac{\partial \Phi_1}{\partial x_2})dt + a_4dB_2(t).
    \end{split}
\end{equation}

\begin{figure}[ht]
\centering
\includegraphics[width=1\linewidth]{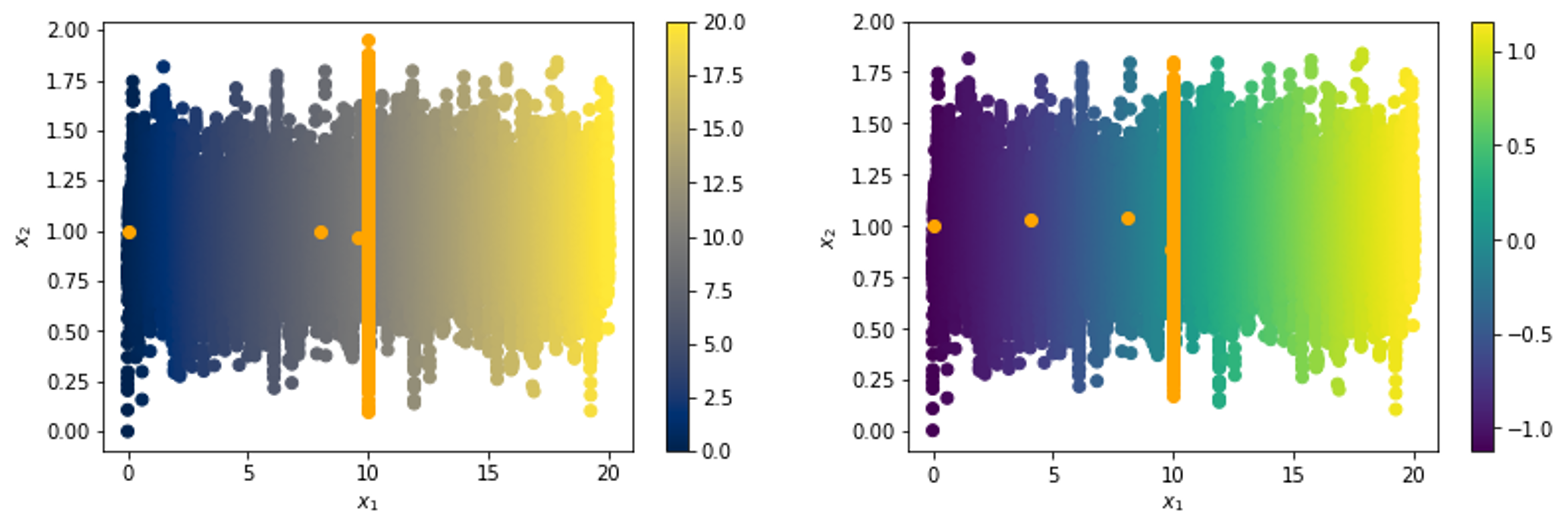}
\caption{US outputs (orange points) obtained by biasing on the known slow variable, $x_1$ (left); and based on the learned diffusion map coordinate, $\phi$ (right), conditioned at $x_1=10$, and the corresponding value in diffusion map space, respectively. The SDE data in the left and right plots are colored by $x_1$ and $\phi$, respectively. Notice the few discrete orange points at the very onset of the US computation, starting from the initial condition (0,1) and quickly approaching
the conditional equilibrium measure.}
\label{fig:US_on_dmaps}
\end{figure}

Results in Figure \ref{fig:US_on_dmaps} show that the US approach of adding a harmonic biasing potential to steer dynamical systems described by SDEs can be performed
either in physical space (if the slow variable is known) or in learned coordinates. 
Work using diffusion map coordinates to guide or direct molecular dynamics simulations \cite{imapd, dm-d-md, Laing_2007} or linking them with Monte Carlo techniques \cite{FergusonDmapColloids} has been demonstrated, but US work biasing conditioned on a diffusion map coordinate value (including calculations of the requisite gradients) has not, to the best of our knowledge, been published before. 
\section{Conditional generative adversarial network sampling}
\label{sec:cGAN_sampling}
A primary goal of the biased simulations in US is to
use sampling of the biased distribution (as an intermediate step) to eventually extract effective sampling of the equilibrium distribution \emph{conditioned} on the values of some order parameters of the system (\emph{e.g.,} dihedral angles, principal components, etc.). 
Such sampling can also be accomplished effectively through the use of a GAN. A generative adversarial network \cite{goodfellow2014} consists of two neural networks: the generator and the discriminator. The generator takes in $n$-dimensional noise and is trained to produce a target output, while the discriminator is trained to distinguish true data from the generator's output. The training is performed in an iterative fashion to optimize a given loss function (often the minimax loss \cite{goodfellow2014} or Wasserstein loss \cite{wgan}). Conditional GANs \cite{mirza2014} operate with the same procedure as a vanilla GAN, but have an additional input to the generator in the form of a label. The GAN then generates conditional distributions of reasonable data given the prescribed label provided to it. We can use this cGAN architecture (given training data from the dynamical system) to act as a data-driven tool for ``lifting.'' A cGAN was trained on a data set directly produced by the system of SDEs given in equation (\ref{eq:untransformed_SDE}), with the conditional training labels being the values of the slow variable at each point. Note that cGANs in literature have mainly been applied to classification problems and have thus had labels based on discrete categories. The multiscale dynamical systems of interest in this work all have continuous CVs, so we use a subclass of cGANs, the Continuous conditional GAN (CcGAN) \cite{ding2021ccgan}, that has an architecture and loss function explicitly constructed to condition on continuous labels. A brief overview of the CcGAN architecture, loss function, and parameters used in this work are detailed in Appendix \ref{sec:app-b}. Figure \ref{fig:US_vs_GAN_hist} shows the data generated by a cGAN conditioned on the label $x_1=10$ compared with the output of US obtained through the process detailed in section \ref{sec:US_on_dmap_CVS}. Again note (as first mentioned briefly in section \ref{sec:US_on_dmap_CVS}) that when performing US, it becomes necessary to remove the artificial biasing force in order to produce the true slow variable distributions of the system, or in other words, ``reweight'' the biased distribution back to the unbiased joint distribution (this is included in the previously mentioned WHAM \cite{wham} and MBAR \cite{mbar} workflows, among others). In this work, we are only focused on generating conditional distributions of the fast variables, so reweighting to find the distribution of the slow variable is not necessary. To further emphasize this point in this first toy example, $x_1$ and $x_2$ are orthogonal, so $x_1$ and $x_2$'s conditional distributions are completely independent.
\begin{figure}[ht]
\centering
\includegraphics[width=1\linewidth]{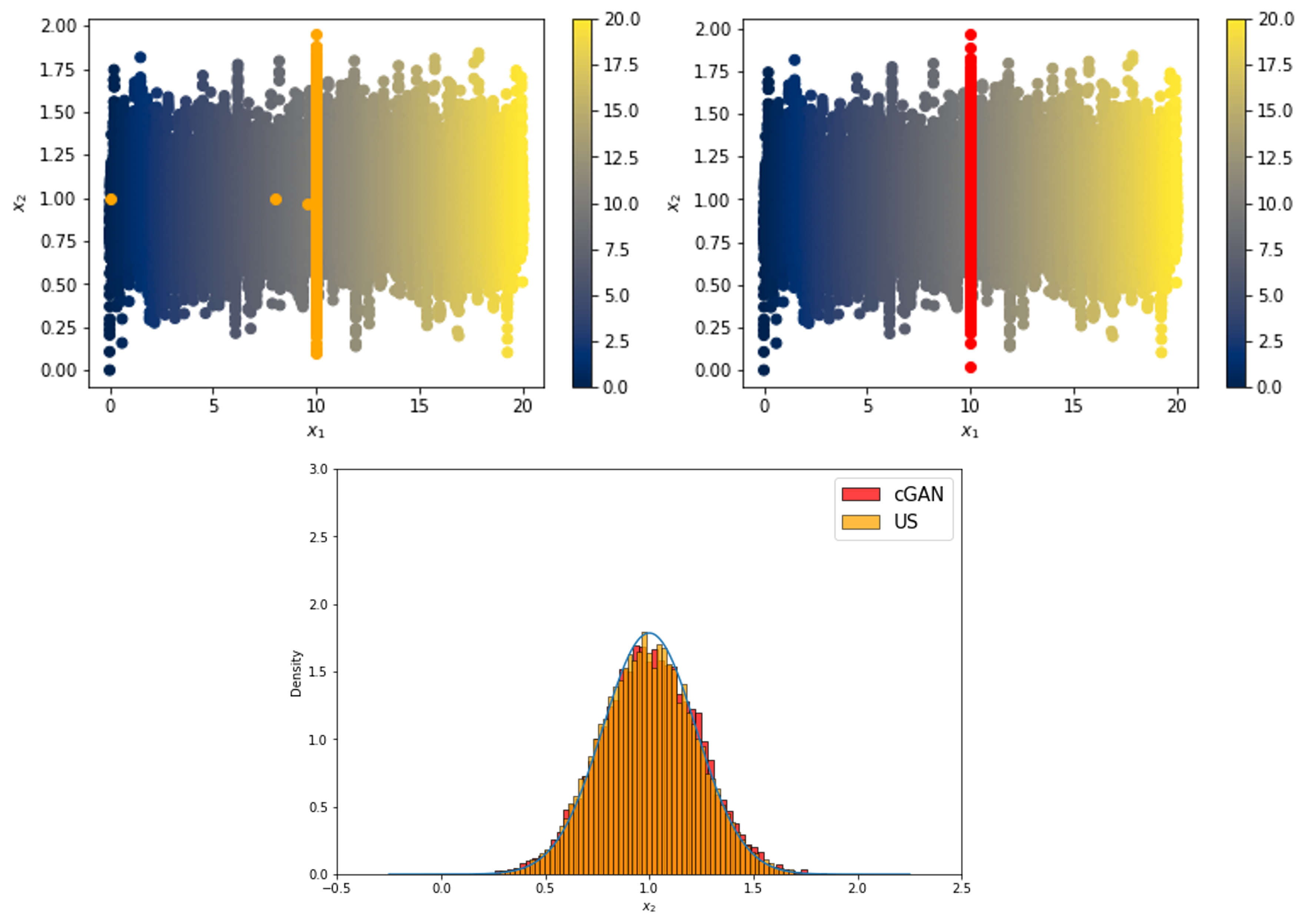}
\caption{US (orange, top left) and cGAN (red, top right) outputs shown for the value $x_1=10$ of the conditioned slow variable, with histograms (bottom) of the $x_2$ values of the outputs compared with the true conditional probability distribution function, shown as a light blue line. The SDE data in the plots are colored by the \emph{a priori} known slow variable, $x_1$.}
\label{fig:US_vs_GAN_hist}
\end{figure}
Figure \ref{fig:US_vs_GAN_hist} shows that conditional GANs trained based on data for this dynamical system can produce fine scale data consistent with samples of the desired conditional probability distribution. 
It is worth noting that US requires an initial condition and must take ``a few'' initial steps as it approaches the appropriate conditional distribution; in contrast to this, the cGAN, once trained, can immediately produce samples from the (approximated) conditional distribution without initial equilibration.
After training, the cGAN and US outputs are compared against the true conditional distribution, a Gaussian distribution with mean one and variance $0.05$. We restate that this very simple illustrative example was performed with the \emph{a priori} knowledge of the slow direction, which, to boot, was orthogonal to the fast direction. Utilizing diffusion maps (instead of the known slow variable $x_1$) with US (or cGANs, as shown in this section) will also produce equivalent data, as discussed/shown in section \ref{sec:US_on_dmap_CVS}.

\section{GAN assisted physics-based sampling}
\label{sec:gan-assisted}

After illustrating (through a simple caricature) the capabilities of the two approaches towards lifting (the physics based US and the data/ML driven GANs) and pointing out some advantages and disadvantages of the two techniques, we now proceed to link US and cGANs together towards creating a more computationally efficient sampling framework.

\subsection{A less trivial multiscale system}

Consider a new system of coupled SDEs in which the slow variable {\em is not available in advance} for biasing. 
We obtained this by applying a nonlinear transformation to the previous simple system described by Equation (\ref{eq:untransformed_SDE});
the simulation dynamics now take the form of ``half-moons'' evolving radially. The transformation is:
\begin{equation*}
    \begin{split}
        &y_1 = x_2\cos{(x_1+x_2-1)},\\
        &y_2 = x_2\sin{(x_1+x_2-1)}.
    \end{split}
\end{equation*}
The slow variable must now be found through data mining. Here again we use diffusion maps, but with an alternative distance metric to the standard Euclidean distance: the {\em Mahalanobis distance}. The Mahalanobis distance measures normalized distances with respect to the variances of each local ``principal direction'' of the noise, allowing us to retain information describing both the fast and slow directions of the system \cite{mahalanobis1936generalized, SINGER_coifman_dmaps, DsilvaSDE}. The Mahalanobis distance, as originally defined, is for a constant or fixed covariance, so for our use (which has a covariance that changes based on the system's position in $y_1$ and $y_2$), we will use a modified definition proposed by Singer and Coifman \cite{SINGER_coifman_dmaps}. 

\begin{figure}[ht]
\centering
\includegraphics[width=1\linewidth]{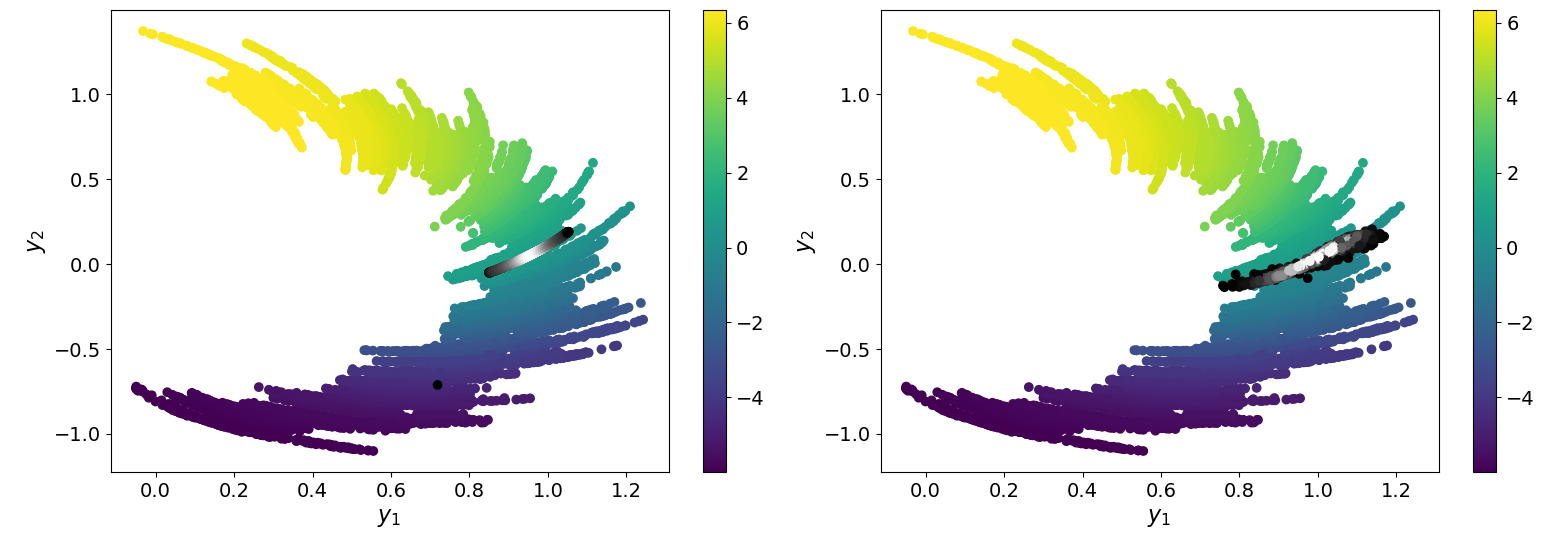}
\caption{Training data obtained from the transformed SDEs colored by the learned slow direction, $\phi$ US (left) and cGAN (right) outputs shown at the conditioned variable $\phi=0$ and shaded by density. White is the highest density, while black is the lowest density.}
\label{fig:Erban_density_comp}
\end{figure}

This modified Mahalanobis distance between two points $x_i$ and $x_j$ can therefore be written as
\begin{equation*}
  \norm{x_j - x_i}^2_M=\frac{1}{2}(x_j - x_i)^T\left( C^\dagger(x_i) + C^\dagger(x_j) \right)(x_j-x_i),
\end{equation*}
where $C(x_i)$ is the covariance of the noise at the point $x_i$ and $\dagger$ signifies the Moore--Penrose pseudoinverse.
This metric will prove useful when the noise contains strong correlations, as is the case in our example since $y_1$ and $y_2$ contain information from both the slow $x_1$ and the fast $x_2$ directions.
Using the tools and algorithms detailed in sections \ref{sec:US_on_dmap_CVS} and \ref{sec:cGAN_sampling} above, but on the transformed data now, we train a cGAN to generate samples from the conditional equilibrium measures.
US also accomplishes this, but a further step is required before we run biased simulations: the biasing potential cannot be simply added to the original equations and then transformed. It must be added directly to the transformed equations post-transformation. This is due to the fact that if the bias is added to the pre-transformed equations and subsequently transformed, it will not have the same behavior as the pre-transformed bias and will often explode or bias toward some undesired location in the transformed space.
Thus, It\^o's lemma \cite{ito} must be applied to transform the original equations to the new system that we will bias. Application of the lemma and the equations that result (and are then subsequently biased) can be found in Appendix \ref{sec:app-e}.
The effective variable is now the leading diffusion map coordinate, $\phi$ (that is, incidentally, one-to-one with the original slow direction $x_1$; see and compare Figures \ref{fig:US_vs_GAN_hist} and \ref{fig:Erban_density_comp}). Physics-based and GAN-based sampling are performed in terms of this diffusion map coordinate.
\subsection{Linking a cGAN with US and its benefits}

\begin{figure}[ht]
    \centering
    \includegraphics[width=0.6\linewidth]{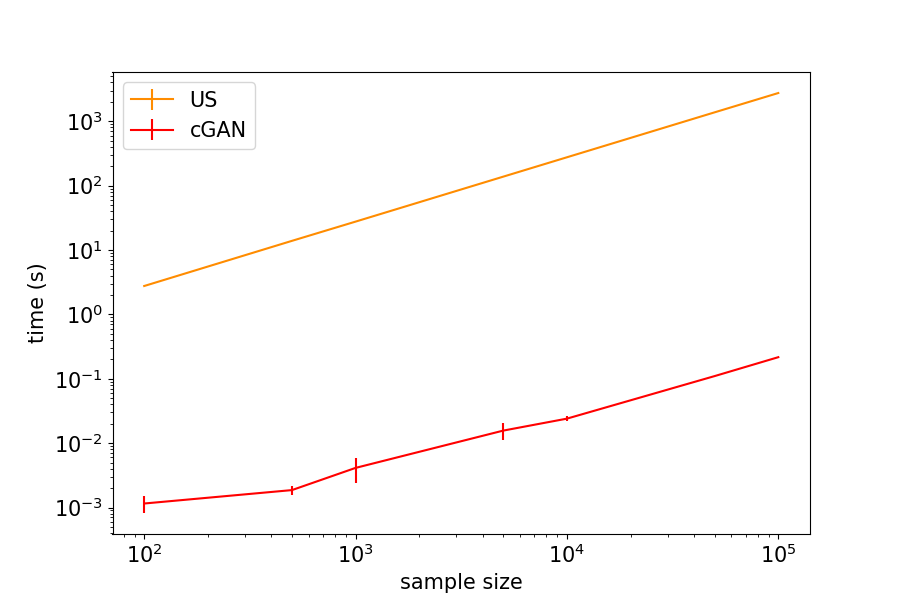}
    \caption{A logarithm-scale plot with error bars of the time required to produce conditional data of increasing sample sizes for US (orange) and a cGAN (red).}
    \label{fig:cost}
\end{figure}

In our transformed setting, both the cGAN and the US (possibly assisted by WHAM) produce acceptable ``liftings''/samples from the conditional equilibrium distributions.
The computational cost associated with each method, however, is very different. A comparison of computational time/cost appears in Figure \ref{fig:cost}: the time required for each method to compute varying sample sizes from the targeted conditional distribution.
Due to the US requirement for physical dynamics (integration of the SDEs), US takes multiple orders of magnitude longer to produce the relevant data, in addition to requiring initialization (less stringent) and an initial ``warmup'' phase, as discussed in section \ref{sec:cGAN_sampling}. 
The cGAN produces the conditional data significantly faster; however, collection of the training data set and training of the networks are required --offline-- before the cGAN can start producing these conditional distributions. 
The training of the GAN networks is essentially \emph{an up-front computational cost}, which is paid to enable later rapid production of reasonable conditional data. 
While this initial cost is significant, the GAN circumvents the cost of propagating the dynamical system forward in time once it is trained. 
The savings in computational time and cost become more evident with increasing sample sizes and increasing complexity of the dynamical system under investigation. 
The cGAN of course also requires initial unbiased simulation/observations to put together the training data set. The training data for the dynamical system detailed in this section consisted of 30,000 samples produced in an unbiased run spanning approximately 6 minutes (360 seconds) of wall time; the information about the platform can be found in Appendix \ref{sec:app-d}. The average time taken to train the cGAN, as detailed in Appendix \ref{sec:app-b}, is approximately 21 minutes (1300 seconds), which is approximately equivalent to the time it takes on the same platform for US to integrate the forces of the system and propagate in time for 100,000 time-steps (and thus, producing 100,000 samples of conditional data since we sample the data at every time-step). Note that 100,000 time-steps is more than enough for US to equilibrate and fully span the desired equilibrium measure.

Once trained, the cGAN can sample 100,000 times from a prescribed conditional distribution in 0.3 seconds (again, on the same platform used in previous benchmarking). 

\begin{figure}[ht]
    \centering
    \includegraphics[width=0.6\linewidth]{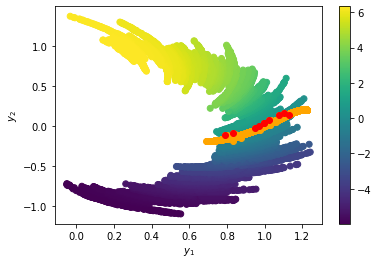}
    \caption{Ten initial conditions generated by a cGAN (red points) were used to initialize 10 parallel US runs, each 1000 time steps 
    long, that clearly visually sample the conditional equilibrium measure.}

    \label{fig:coupling}
\end{figure}

Furthermore, in order to obtain the true distribution of the variable in which the condition/force is added and subsequently the potential of mean force, the added bias from US must be removed (through WHAM, MBAR, etc.). 
The cGAN-generated conditional distribution requires no removal of a biasing term (but it implicitly assumes that a normally distributed seed for the cGAN samples the conditional equilibrium distribution accurately). 
Clearly, the trained cGAN is a much more ``real time'' cost efficient option, once the up-front cost of data generation and training is expended. Yet, while it quickly and efficiently samples the conditional equilibrium measure, it is ultimately an approximation of the ground truth based on how representative the collected training data set was.
It appears reasonable to ``accentuate the positive, eliminate the negative'' of each of the two approaches (the negatives being long times for US simulations, possibly inaccurate data-based approximation for the cGAN) by combining them: we suggest that it is possible to initialize many parallel runs of US with ``warmed up'' initial conditions constructed from the output of the trained cGAN. Figure \ref{fig:coupling} displays an output of multiple parallel US runs initialized by a trained cGAN.

While both approaches of US alone and US coupled with a cGAN initialization are capable of spanning the desired full equilibrium measure, running ten parallel biased sampling simulations of 1,000 time-steps initialized by cGAN outputs results in accurate conditional distribution sampling that is obtained ten times faster (in wall-clock time) than a single, long, biased US simulation of 10,000 time-steps. 
The data (conformations) produced now arise directly from the dynamical system simulation (and may thus be more plausible than cGAN approximate constructs).

\subsection{Analysis of coupling accuracy and convergence}
\begin{figure}[ht]
\centering
\includegraphics[width=1\linewidth]{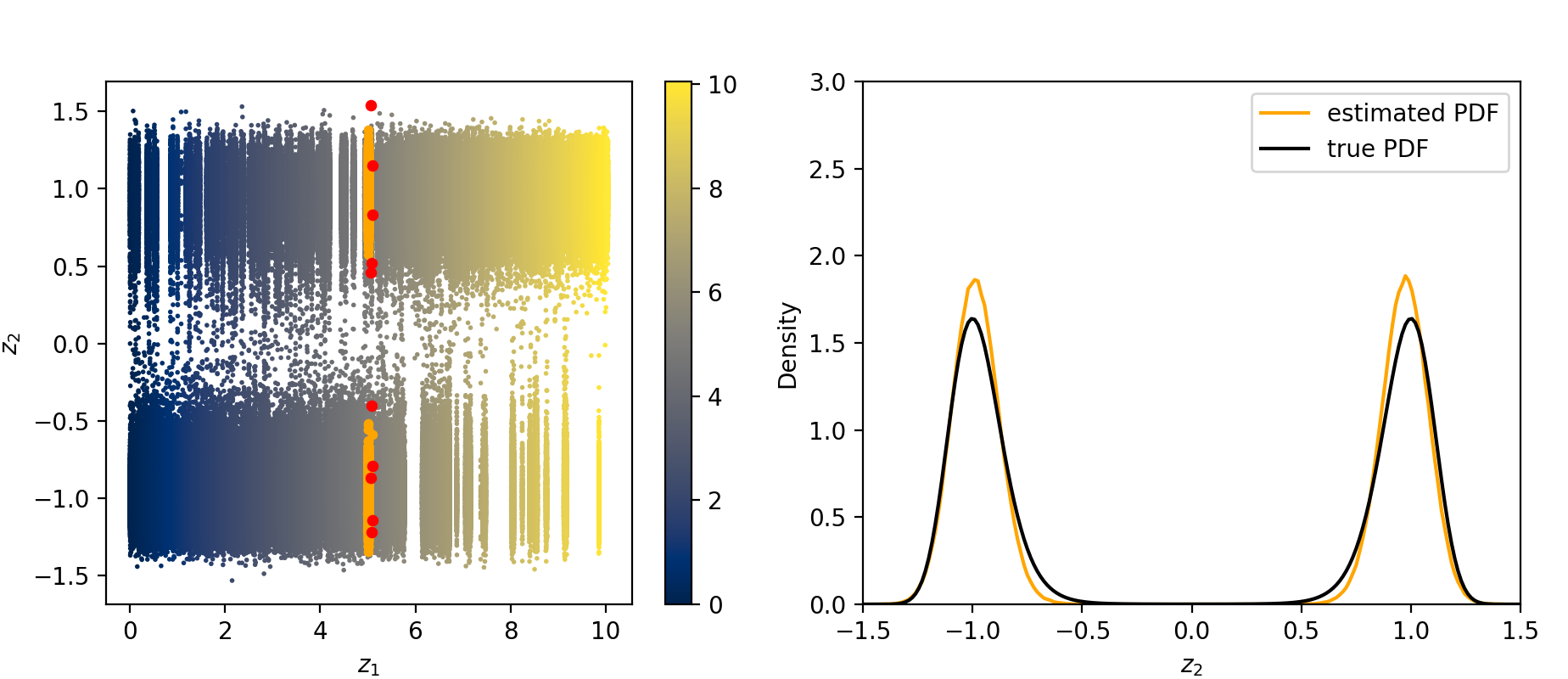}
\caption{On the left, we show ten initial conditions generated by a cGAN (red points).
  Initial conditions drawn from the cGAN such as those shown above were used to initialize ten parallel US runs (orange points) at $z_1 = 5$, where the free energy basins are at equal depth (additionally, $h = 8$). Each US run is $10^4$ time steps long, and when combined they sample the conditional stationary probability. On the right, the probability density function for $z_2$ at $z_1 = 5$ (orange line), estimated from the cGAN+US coupled runs pooled together, is compared with the true probability density function of $z_2$ at $z_1 = 5$ given by the known free energy function (black line).}
  \label{fig:dwell_evolving_pdf}
\end{figure}

\begin{figure}[ht]
    \centering
    \includegraphics[width=1\linewidth]{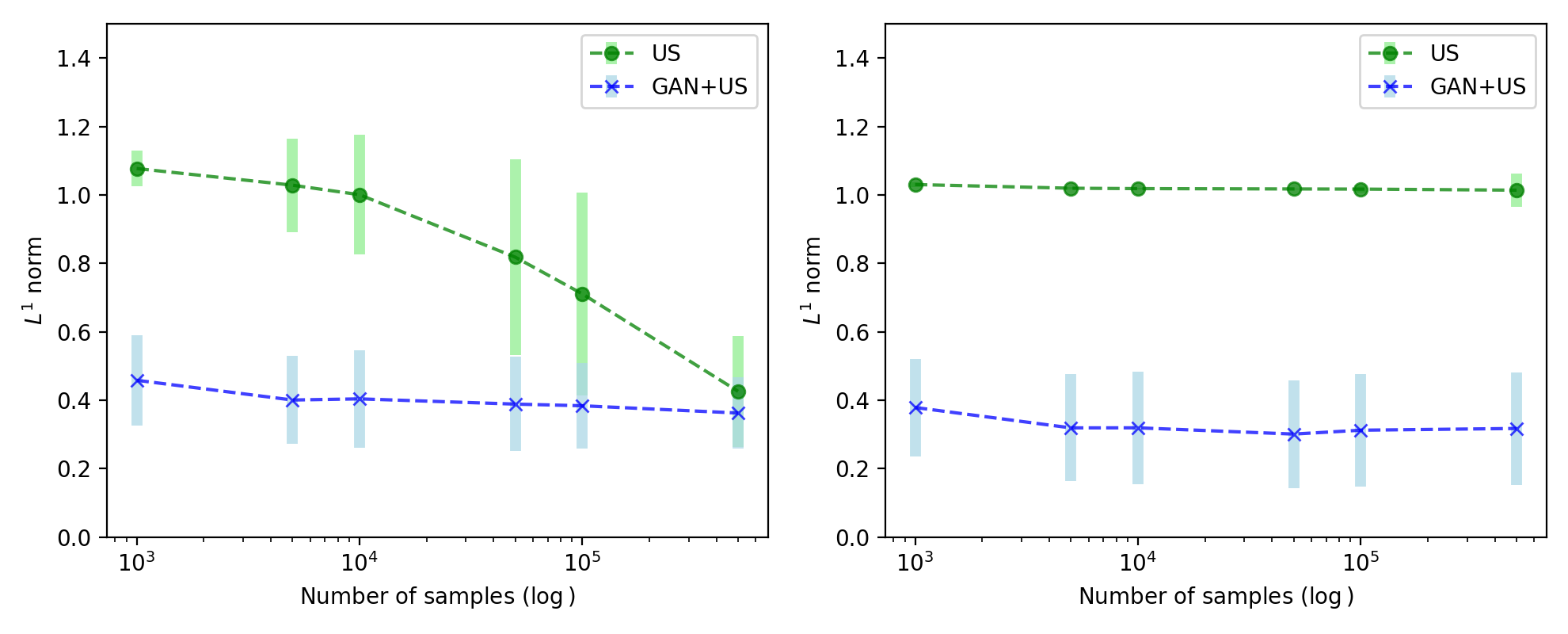}
    \caption{Experiments run at two different barrier heights ($h=4$ for the left plot, $h=8$ for the right plot) showing the $L^1$ norm of the absolute value of the difference between the true PDF and the estimated PDF determined by US (green) and coupled cGAN+US (blue) runs.
      The $L^1$ norm is a function of the sample size (\emph{i.e.,} number of time steps).
      Each data point and its corresponding error bar reflect the results of 1,000 independent experiments.
      Note that the $L^1$ norm error is high with small variance for the US runs when $h=8$ due to US alone not being able to cross over the high free energy barrier and sampling only a single basin out of the existing two.}
    \label{fig:dwell_cost}
\end{figure}

 In order to investigate improving efficiency by coupling GANs with US, we consider an example that exhibits metastability in the fast direction, which results in a non-Gaussian distribution of samples.  Consider the new system:
{\scriptsize
\begin{equation}
    \begin{split}
    dz_1(t) &= a_1 dt + a_2 dB_1(t),\\
    dz_2(t) &= -((1 + z_2(t))^2 (2 (1 + h - (0.4 z_1(t) - 2))z_2(t) - 2 h + 3 (0.75 (0.4 z_1(t)          - 2) - 2) z_2(t)^2 \\  
            &+ 4 z_2(t)^3) + 2 (1 + z_2(t)) (h - 2 h z_2(t) + (1 + h - (0.4 z_1(t) - 2)) z_2(t)^2 \\ &+ (0.75 (0.4 z_1(t) - 2) - 2) z_2(t)^3 + z_2(t)^4))dt + a_3 dB_2(t).
    \end{split}
    \label{eq:dwell}
\end{equation}
}
where $B_1$ and $B_2$ are standard Brownian motions, $a_1 = a_2 = 10^{-4}$, and $a_3 = \epsilon{a_1} = \epsilon{a_2} = 10^{-1}$ where the ratio of time scales is $\epsilon = 10^3$;
these scalings imply that $z_1$ is the slow variable and $z_2$ is the fast variable ($z_2$ is faster than $z_1$ by a factor of $10^3$). This system differs from the untransformed system of section \ref{sec:cGAN_sampling} in that the stationary probability in the fast direction is not a Gaussian distribution, but a multimodal distribution with basins centered at $z_2 = 1$ and $z_2 = -1$. The height of the energy barrier is defined by the variable $h$, and the depth $d$ of the well at $z_1 = 1$ is a function of $z_1$. Namely, $d = 0.2 z_1 - 2$. This means that the potential energy (and by extension, the equilibrium distribution) of $z_2$ changes as $z_1$ increases. Knowing the distribution of $z_2$ in a local area of $z_1$ values does not define the distribution of $z_2$ everywhere; yet we can train a cGAN on local values and couple the cGAN generated initial conditions with physics-based biased sampling. We  performed a computational convergence comparison using the number of force evaluations of the system (in other words the number of time steps in which the SDEs are integrated). We ran sets of US simulations and US simulations initialized by cGAN provided initial conditions with increasing sample sizes that directly correspond to the number of times the SDEs are integrated (from $10^3$ to $5 \times 10^5$ samples). We then averaged these simulations and used their data to compare with the \emph{a priori} known free energy surface given by \eqref{eq:dwell}. Figure \ref{fig:dwell_evolving_pdf} shows an example output of the coupling method along with a comparison of its probability density function (PDF) with the true conditional PDF of the system. Figure \ref{fig:dwell_cost} compares the convergence capabilities of the coupling method with the capabilities of US alone.

In this case where the conditional distributions exhibit metastability, the coupling of a cGAN with US initializes US runs from initial points at multiple metastable states and their estimates are pooled together. If there were biases present in the fast direction, the histograms could be reweighted and combined by using WHAM\cite{wham}. This results in a further increased performance when compared to US alone, especially at smaller sample sizes, because US alone tends to become trapped in the metastable states. In addition to the already faster convergence, the multiple runs of the coupled method can be trivially parallelized.

\section{Application to molecular dynamics simulations of alanine dipeptide}
\label{sec:alanine}
Conditional cGAN-based sampling is not limited to dynamical systems described by SDEs. 
It has been demonstrated, for example, that machine learning techniques can also be applied to molecular dynamics data for purposes such as interpolation \cite{mlss} and learning effective dynamics \cite{Vlachas_acceleration_MDN}.
In this section, we demonstrate the use of a cGAN for sampling conditional equilibrium measures of molecular dynamics (MD) simulation data and compare the results directly to established US methods. 
\begin{figure}[ht]
    \centering
    \includegraphics[width=0.75\linewidth]{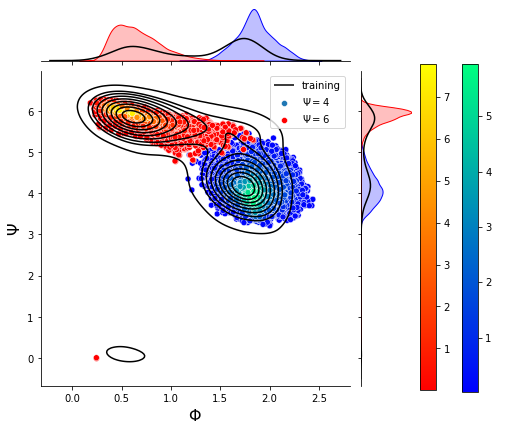}
    \caption{Dihedral angles of many cGAN output samples; red (blue) corresponds to conformations produced by a cGAN conditioned on
    the $\Psi$ value of the left (right) free energy minimum. 
    The sole point at the bottom right of the plot arises from the $2\pi$ periodicity of the $\Psi$ angle.
    Color gradients (red to yellow, blue to green) represent empirical sample densities. The continuous contours represent the empirical sample density of a long unbiased simulation run. 
    The empirical marginals for the cGAN sampling (in color) and for the data collection run (as black curves) are included to guide the eye.}
    \label{fig:ala46}
\end{figure}

\begin{figure}[ht]
\centering
\includegraphics[width=1\linewidth]{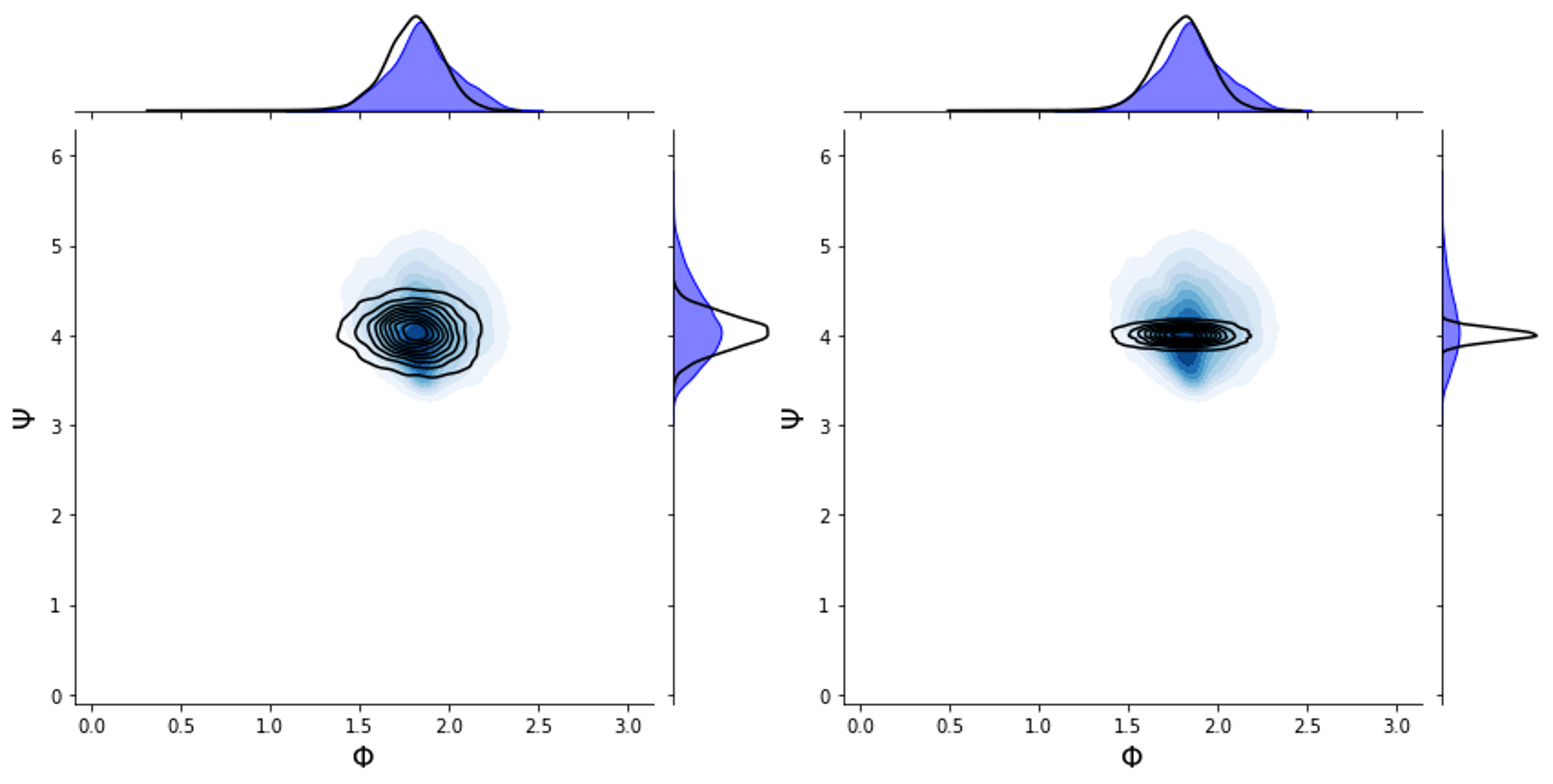}
\caption{$\Phi - \Psi$ distribution of cGAN outputs conditioned on $\Psi=4$ (blue), shaded by empirical density, compared with the empirical density of US biased simulation data (black contour lines) at two different values of $\kappa$ before WHAM reweighting: $\kappa = 50$ kJ/mol (left) and $\kappa = 500$ kJ/mol (right). The empirical marginals of the  US and cGAN outputs with respect to $\Phi, \Psi$ are shown on the margins.}
\label{fig:ala_kappa_comp}
\end{figure}

The training data for the cGAN was produced by a 20 nanosecond MD simulation of alanine dipeptide in implicit solvent
that sampled two local free energy basins at distinct values of $\Phi$ and $\Psi$ (the {\em a priori} established pair of collective variables, dihedral angles that summarize the bulk behavior of the molecule \cite{ala_minima}). Further simulation details are discussed in Appendix \ref{sec:app-d}. 
US biasing can be performed using a single CV or even several CVs \cite{TORRIE1977}. Here, the potential wells are distinct in both $\Phi$ and $\Psi$, so either variable (or both variables) could be an appropriate choice on which to bias. For our demonstration, we choose here to bias on $\Psi$. 
A cGAN was trained on the collected data to produce molecular configurations conditioned on $\Psi$ values that correspond to each one of the two basins. The data produced are 22-dimensional molecular conformations, but in Figures \ref{fig:ala46} and \ref{fig:ala_kappa_comp} they are represented based on their $\Phi$ and $\Psi$ angles. 
One could also condition on data-driven, diffusion map coordinates, learned (as detailed in section \ref{sec:US_on_dmap_CVS}) from processing the training data set.
%
Translational and rotational variances were factored out in the training data through the Kabsch algorithm \cite{Kabsch:a12999}. The cGAN architecture and training protocol are described in Appendix \ref{sec:app-b}.
During training, all generator output conformations were aligned to a reference conformation of alanine dipeptide using the Kabsch algorithm before the loss function was computed.

The cGAN produces molecular conformations that are consistent with the prescribed label (the conditioning variable); furthermore, the cGAN learns to produce conditional distributions whose empirical densities appear
to qualitatively mimic the training data empirical densities. To explore this point further, cGAN results are compared in the left plot of Figure \ref{fig:ala_kappa_comp} with US simulations of alanine dipeptide biased on $\Psi = 4$ corresponding to one of the two wells (the spring constant, $\kappa$, of the US bias was set to 50 kJ/mol). As mentioned in section \ref{sec:US_on_dmap_CVS}, US's distribution of the conditioned variable is dependent on the strength of the force constant (denoted as $\kappa$ in this example). Thus, it is important for the nonconditioned variable distributions to match (here, $\Phi$) while the conditioned variable distribution (here, $\Psi$) is allowed to change.

The cGAN trained on Alanine Dipeptide, when conditioned on one of the two minima, produces output conformation distributions that bear qualitative resemblance to both (a) the unbiased empirical distribution of the training data set for the basin in question and (b) to the output distribution of US biased on the minimum of said basin. 
Umbrella sampling does, of course, provide finer control through its ``hyperparameters'' such as the strength and curvature of the
biasing potential. 
cGAN output distributions are compared in the right plot of Figure \ref{fig:ala_kappa_comp} to a US output where the kappa was ten times stronger than the previous value (500 kJ/mol). The US bias is again only added to $\Psi$, and the results were compared with a CcGAN output conditioned at the same value of $\Psi$.

The US output of course has a more narrow distribution in $\Psi$; the GAN output has no ``adjustment knob'' to fine-tune the conditional distribution it produces. Clearly, however, the coupling approach we proposed allows the cGAN to accelerate US sampling through good --and well-distributed-- initial conditions, so that fine-tuned results can ultimately be collected with improved wall-clock computational efficiency.
\section{Further capabilities of cGANs trained on dynamical systems}
\label{sec:dim_and_extrap}
\subsection{The dimensionality of input noise and its effect on cGAN outputs}
The input to the generator network of a GAN (and cGAN) is typically an $n$-dimensional vector of Gaussian noise. For the dynamical system detailed in section \ref{sec:US_on_dmap_CVS}, the equilibrium measure we desire to reproduce is one-dimensional, so a single input noise seed suffices. 
However, in higher-dimensional dynamical systems, the dimensionality of the noise seeds fed to the generator can and will alter the equilibrium measure learned, and subsequently reproduced, by the neural network model. \par
Consider the dynamical system described by the three-dimensional system of coupled SDEs below:
\begin{equation}
    \begin{split}
    &dx_1(t) = a_1dt,\\
    &dx_2(t) = a_2dt + a_3dB_2(t),\\
    &dx_3(t) = a_4dt + a_5dB_3(t),
    \end{split}
\end{equation}
where $B_2$ and $B_3$ are standard Brownian motions, $a_1, a_2, a_4 = 1, 10, 100$, respectively, and $a_3 = a_5 = \sqrt{0.02}$, meaning $x_1$ is a slow variable, $x_2$ is a fast variable, and $x_3$ is an even faster variable. Now, apply the following transformation to the system:
\begin{equation*}
    \begin{split}
        &y_1 = x_1 + x_2^2 + x_3^2,\\
        &y_2 = x_2^2 + x_3^2,\\
        &y_3 = x_3.
    \end{split}
\end{equation*}
This nonlinear transformation produces evolving ``caps'' (rather than ``half-moons'') along the slow direction, essentially having a two-dimensional fast equilibrium measure in $y_2$ and $y_3$ that evolves along the $y_1$ axis. 
Using such a dynamical system to produce three-dimensional training data, with a two-dimensional fast equilibrium measure, we can investigate the results of feeding noise seeds of different dimensionality to the generator network. 
The conditional GAN labels used in this case are the values of the \emph{a priori} known slow variable $x_1$, but diffusion maps could also be used as detailed in section \ref{sec:US_on_dmap_CVS}. 
\begin{figure}[ht]
\centering
\includegraphics[width=1\linewidth]{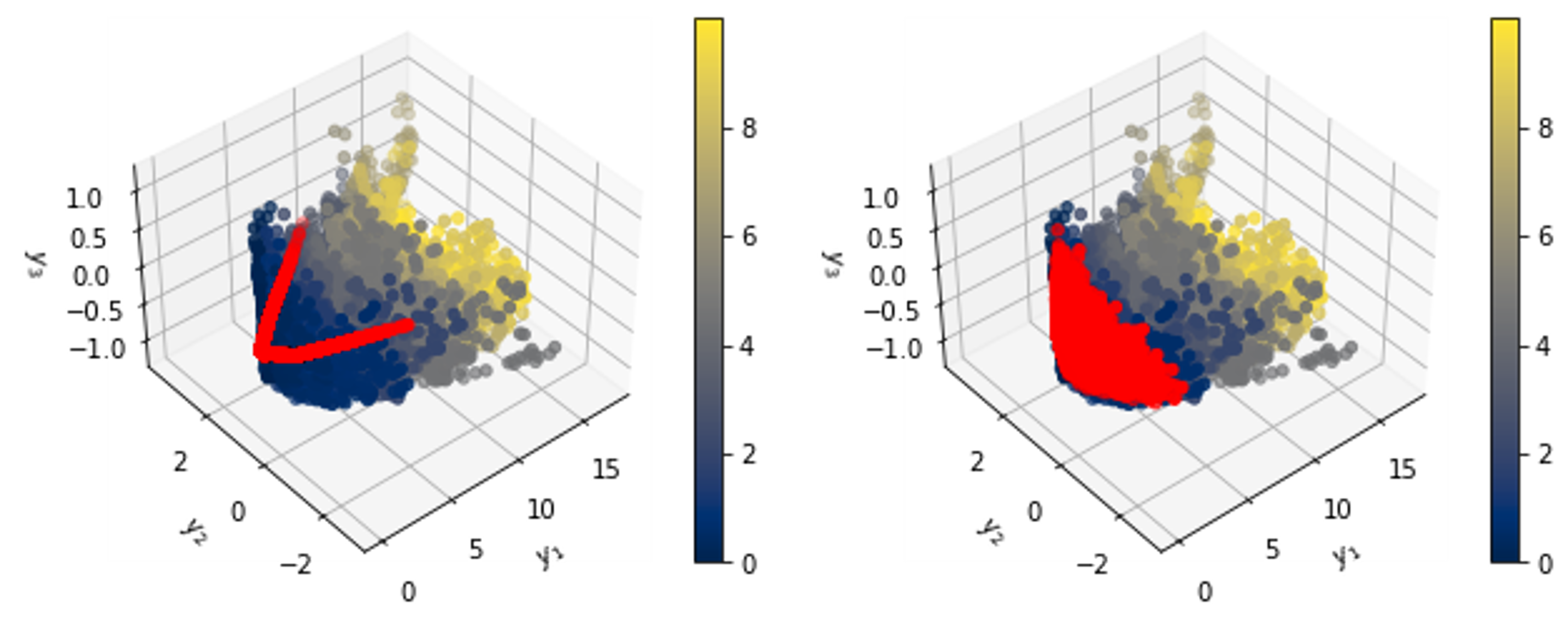}
\caption{cGAN outputs (red) for the value of the conditioning variable $x_1=0$ with 1D input seed noise (left) and 2D input seed noise (right). Here the slow variable $x_1$ is known \emph{a priori}; the training data (a sequence of progressing ``spherical caps,'' colored by $x_1$), form the backdrop. }
\label{fig:caps_noise_dim}
\end{figure}
Figure \ref{fig:caps_noise_dim} displays the resulting cGAN outputs from 1D and 2D generator noise inputs. Given 1D noise, the cGAN produces an ``arc'' along the conditional ``cap'' at the prescribed $x_1$ value; given 2D noise, the cGAN learns to produce the entire conditional measure --the entire ``cap.'' 
This can be significant because in a system with multidimensional equilibrium measures, the cGAN can produce data that is either consistent with the entire conditional measure or consistent with some lower-dimensional subset of the large-dimensional measure, depending on the input fed to the network.
Clearly, the calculation of any statistics based on the cGAN output distribution (e.g., a potential of mean force calculation) will be critically affected. 

\subsection{Towards multiscale evolution: conditional GAN extrapolation}

Consider again the dynamical system detailed in section 7.1. Figure \ref{fig:caps_noise_dim} shows the cGAN can produce a reasonable conditional equilibrium measure at $x_1=0$, and Figure \ref{fig:caps_noise_dim} also shows that the training data ranges from $x_1=0$ to $x_1=10$. The cGAN clearly performs well when producing conditional data within the entire range of the training data; but what if we wish to condition on a variable that is beyond the local range of the training data?

\begin{figure}[ht]
\centering
\includegraphics[width=1\linewidth]{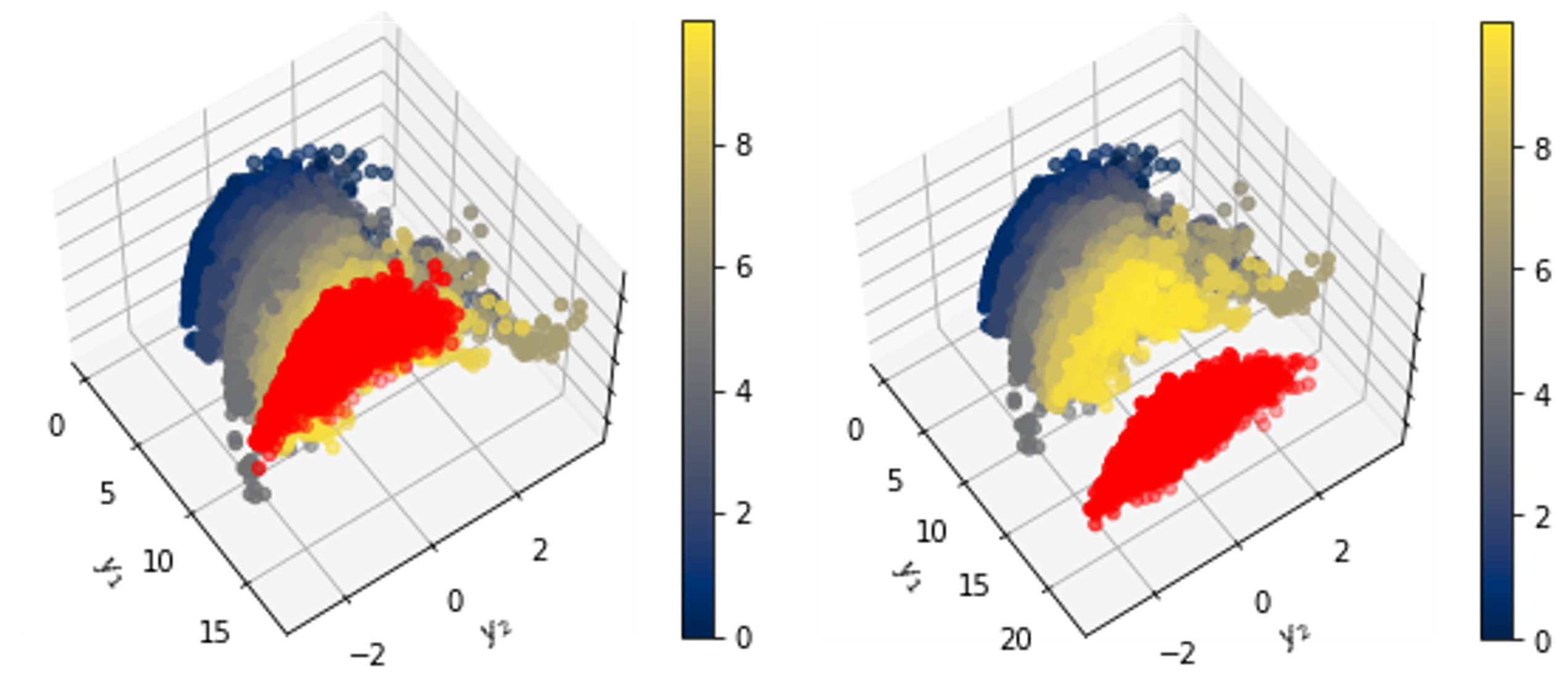}
\caption{cGAN outputs (red), both with 2D input noise, shown at the conditioned variable $x_1=10$ and the extrapolated value of $x_1=20$ where the slow variable $x_1$ is known \emph{a priori}.}
\label{fig:caps_extrapolation}
\end{figure}

Figure \ref{fig:caps_extrapolation} shows that the cGAN can indeed construct a (visually) reasonable initial estimate of the conditional equilibrium measure at $x_1=20$, a conditioning value that is well outside the range of the training data. 
%
%
While the true data at $x_1=20$ is not at our disposal (yet!), our proposed coupling of the cGAN computational technology with an enhanced sampling technique, like US, demonstrated in section 4.2, could expedite the exploration of the phase space of the dynamical system beyond what is contained in the cGAN's training data set.
In this fashion, the cGAN can lead to useful ``lifting'' in equation-free simulations of multiscale dynamical systems.
Providing good, fine scale initial conditions, consistent with extrapolated slow variable values, is a critical step in the implementation of the brief bursts of fine scale simulation
on which bridging fast and slow time scales relies \cite{eqfree, imapd, georgiou2017exploration}.

Our approach is predicated on the diffusion maps algorithm identifying the slow and fast variables and on the application of the averaging principle for stochastic differential equations~\cite{khasminskii1968,freidlin2012} to eliminate the fast variables and close the effective equations on the slow variables.

For instance, consider the following stochastic fast/slow system given in diffusion map coordinates $(\phi_1, \phi_2)$ by
\begin{equation*}
  \left\{
    \begin{aligned}
      \d \phi_1 &= \mu_1(\phi_1, \phi_2) \, \d t + \sigma_1(\phi_1, \phi_2) \d B_1, \\
      \d \phi_2 &= \epsilon^{-1} \mu_2(\phi_1, \phi_2) \, \d t + \epsilon^{-1/2} \sigma_2(\phi_1, \phi_2) \d B_2,
    \end{aligned}
  \right.
\end{equation*}
where $\epsilon > 0$ is the time scale separation parameter, $\mu = (\mu_1, \mu_2)$ is the drift, $\sigma = \diag(\sigma_1, \sigma_2)$ is the volatility, and $(B_1, B_2)$ are components of a standard multidimensional Brownian motion.
If the dynamics on $\phi_2$ conditioned on $\{ \phi_1 = z \}$ have an equilibrium measure $\nu_z^\epsilon(\d \phi_2)$ and the expectation
\begin{equation*}
  B(z, \epsilon)
  =
  \int \mu_1(z, \phi_2) \nu_z^\epsilon(\d \phi_2)
\end{equation*}
converges in the limit of $\epsilon \to 0$ to
\begin{equation*}
  B(z) = \lim_{\epsilon \to 0} B(z, \epsilon),
\end{equation*}
then we can close a reduced effective ODE in terms of the slow variable by integrating over the conditional equilibrium measure.
We thus use the cGAN's ability to sample the conditional  equilibrium measure to {\em close an effective ODE}
for the slow variable $\phi_1$
\begin{equation*}
  \frac{\d}{\d t}{{\phi}}_1 = B({\phi}_1).
\end{equation*}
We refer the reader to \cite{vanden-eijnden2003,legoll2010,hartmann2020} for the specifics on the conditions under which this construction is valid.

\section{Conclusions}
\label{sec:conclusions}
Conditional GANs hold significant parallels with physics-based sampling methods (such as initialization on slow manifolds \cite{init_slow_man1, init_slow_man2}
and US). Coupling the ML-based and the mathematical model-based approaches holds the promise to reduce the required computation time for producing reliable conditional equilibrium state distributions (and for sampling them). 
This is valuable beyond the field of multiscale stochastic dynamical systems: possible applications also arise in materials science, drug discovery, and other similar computational physics and chemistry fields where complex, multiscale/multiphysics models are {\em effectively modeled} by lower-dimensional stochastic differential equations.
Additionally, the cGAN shows promise in producing initial approximations of {\em extrapolated} conditional equilibrium measures for complex systems for values of the slow variables/reactions coordinates {\em beyond} the range of the training data.
Notably, the cGAN's output can be altered by changes to the dimensionality of the generator network's random seed input.
Knowing the collective variables of an effectively simple 
multiscale system beforehand is not a requirement for our works: biased sampling/cGAN training can be performed directly on coordinates produced by manifold learning/dimensionality reduction techniques such as diffusion maps. 
In future work, we hope to explore these avenues in molecular modeling-based multiscale applications in addition to dynamical systems with increasingly complex conditional distributions. On the mathematical side, it would be important to study the rate of convergence of cGAN-produced conditional distributions to corresponding theoretical or empirical training ones. Comparisons with modern developments in generative models (specifically, diffusion models) in the conditional context will also be pursued.

\appendix
\section{Coupling cGANs with umbrella sampling applied to alanine dipeptide simulations and the original untransformed SDE system}
\label{sec:app-a}
cGAN and US coupling was also performed on the alanine dipeptide example from section \ref{sec:alanine}. For this demonstration, five cGAN generated initial conditions at the dihedral angle value $\Psi = 4$ were used to initialize five 0.2 nanosecond US simulations for the same prescribed value where the kappa value was set to 100 kJ/mol.
\begin{figure}[ht]
\centering
\includegraphics[width=0.8\linewidth]{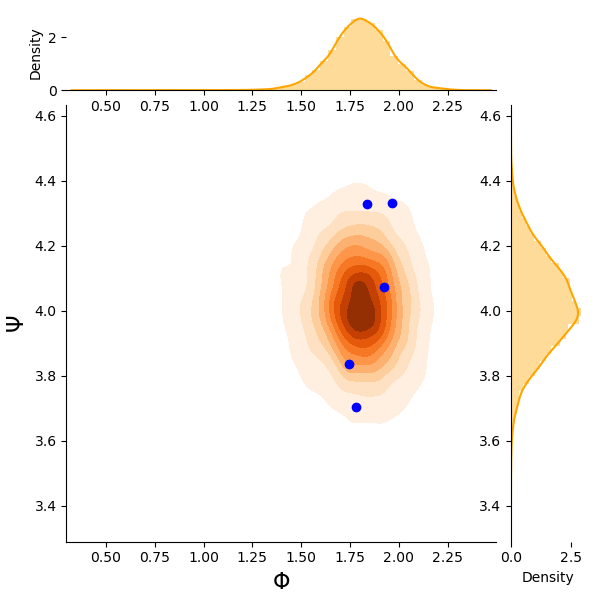}
\caption{Five cGAN generated initial conditions of alanine dipeptide conformations represented by dihedral angle values in blue plotted with the resulting configurations from five US simulations represented with orange shading. The densities for the respective angle values are plotted in the margins.}
\label{fig:ala_coupling}
\end{figure}
In the case of the alanine dipeptide simulations, the cGAN again provides reasonable initial condition configurations, and can expedite the initialization of the simulations. Coupling is not demonstrated with the 3D example from section \ref{sec:dim_and_extrap} due to It\=o's lemma (demonstrated in Appendix \ref{sec:app-e}) not providing accurate equations for the post-transformation system, thus making umbrella sampling impossible.
Additionally, the proposed coupling of cGANs with US was performed with the dynamical systems in the other sections. For example, the dynamical systems described in sections \ref{sec:US_on_dmap_CVS} and \ref{sec:cGAN_sampling} provide additional insight into the capability of this method to reproduce the system's true empirical conditional distributions.
\begin{figure}[ht]
\centering
\includegraphics[width=0.9\linewidth]{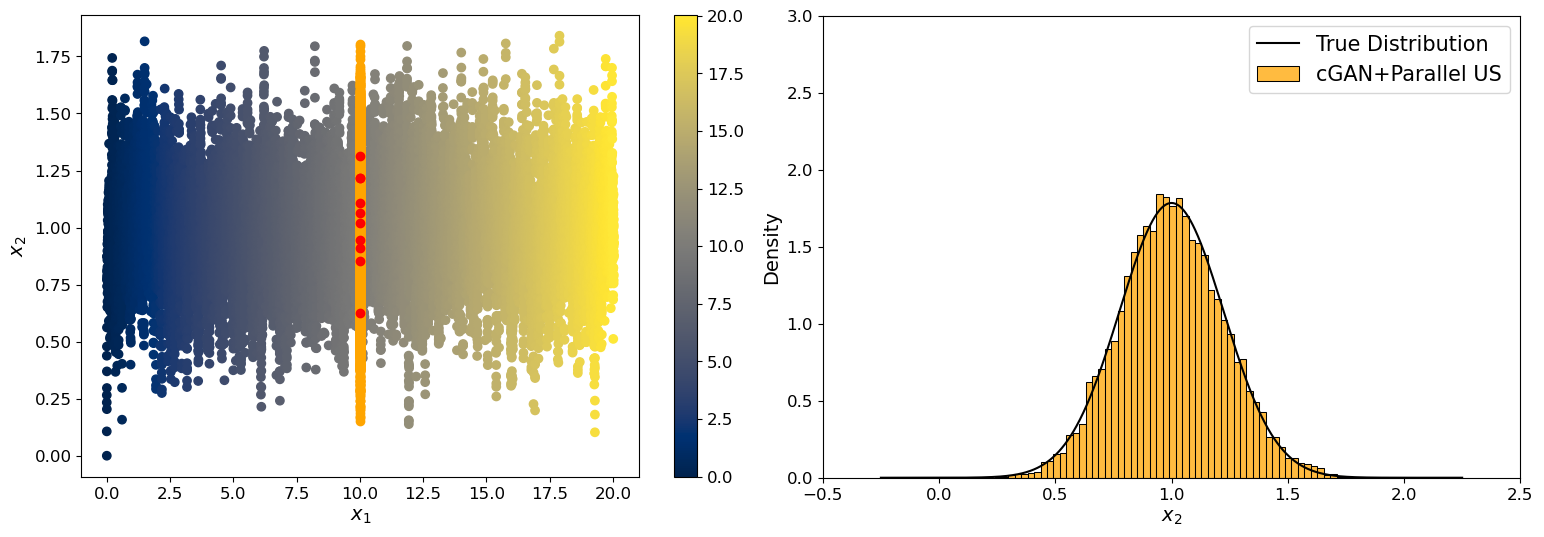}
\caption{On the left, ten initial conditions generated by a cGAN (red points) were used to initialize 10 parallel US runs (orange points), each 1000 time steps long, that clearly visually sample the conditional equilibrium measure. On the right, the $x_2$ values of the data from the US runs are compared with the $x_2$ distribution of the training data (the light blue line).}
\label{fig:untransformed_coupling}
\end{figure}
Like the transformed case in section \ref{sec:gan-assisted}, the US data again is shown to span the fast variable equilibrium measure. This analysis is not performed on the transformed case from Section \ref{sec:gan-assisted} due to the fact that one would need to find a second eigenvector from diffusion maps that is one-to-one with the fast direction. This makes rigorous analysis of the fast variable distribution objectively difficult, but it follows that a system that is transformed by some deterministic equations would maintain the same distribution, albeit transformed. Figure \ref{fig:US_vs_GAN_hist} provides a qualitative comparison of cGAN and US output densities in the transformed space.

\section{CcGAN architecture and training}
\label{sec:app-b}
In the literature, the cGAN model was originally used for classification using discrete labels, so for dynamical systems labeled along a continuous variable, some adjustments are needed. The conditional GAN general architecture , training procedure, and loss function (the \emph{hard vicinal discriminator loss}, specifically) used in this paper were adapted from the CcGAN, proposed by Ding et al. \cite{ding2021ccgan}. The CcGAN's significant difference from the cGAN is the accommodation of using continuous training data labels through the introduction of novel loss functions. The \emph{hard vicinal discriminator loss} (HVDL), which was used for the adapted CcGAN in our work, is shown below: 
\begin{equation*}
    \begin{split}
        \mathcal{L}^{HVDL}(D) = &-\frac{C_1}{N^r}\sum_{j=1}^{N^r}\sum_{i=1}^{N^r}\mathbb{E}_{\epsilon^r\sim\mathcal{N}(0,\sigma^2)} \left[ \frac{\mathbbm{1}_{{|y_j^r+\epsilon^r-y_i^r|\le\kappa}}}{N^r_{y_j^r+\epsilon^r,\kappa}}\log(D(x_i^r, y_j^r + \epsilon^r)) \right]\\
        &-\frac{C_2}{N^g}\sum_{j=1}^{N^g}\sum_{i=1}^{N^g}\mathbb{E}_{\epsilon^g\sim\mathcal{N}(0,\sigma^2)} \left[ \frac{\mathbbm{1}_{{|y_j^g+\epsilon^g-y_i^g|\le\kappa}}}{N^g_{y_j^g+\epsilon^g,\kappa}}\log(1 - D(x_i^g, y_j^g + \epsilon^g)) \right].
    \end{split}
\end{equation*}
The HVDL is a loss function that operates very similarly to the original min-max loss of GANs, except that indicator functions are used to create cutoff distances of acceptable values for the continuous labels. Only the discriminator loss is shown because the generator loss is just the maximization of the generator term in the discriminator loss (the second term where many variables are annotated with a $g$ for generated). $D$ represents the discriminator, and functions of $D$ are the discriminator's output given the variables inside the parenthesis. $N^r$ and $N^g$ are, respectively, the number of real and fake data points, $x_i^r$ and $x_i^g$ are, respectively, the $i$th real data sample and $i$th fake data sample, $y_j^r$ and $y_j^g$ are, respectively, the $j$th real and $j$th fake data label, $\kappa$ is a hyperparameter describing the vicinity of labels in which the loss function will be calculated, $\epsilon^r \triangleq y - y_j^r$, $\epsilon^g \triangleq y - y_j^g$, and $C_1$ and $C_2$ are constants. This HVDL loss functions by training over a conditional distribution of data given by continuous labels in a vicinity around the prescribed label. The CcGAN was trained for 5000 epochs for the case described in section \ref{sec:cGAN_sampling}, 15000 epochs for the case described in section \ref{sec:gan-assisted} and \ref{sec:dim_and_extrap}, and 40000 epochs for the case described in section \ref{sec:alanine}. The Adam optimizer with $\beta_1$ = 0.5 and $\beta_2 = 0.999$ was used with a learning rate of $1 \times 10^{-4}$ and batch size of 512. Tables \ref{tab:untransformed_net}, \ref{tab:transformed_net}, and \ref{tab:alanine_net} describe the architecture of the networks.
\begin{table}[H]
    \centering
    \caption{Network architecture of the generator and discriminator for the example detailed in section \ref{sec:cGAN_sampling}. This is shown in the following format: layer type (output shape); nonlinearities where ``BN'' represents batch normalization. } 
    \begin{tabular}{!{\VRule[2pt]}c!{\VRule}c!{\VRule[2pt]}}
    \specialrule{2pt}{0pt}{0pt}
    Generator & Discriminator\\\specialrule{2pt}{0pt}{0pt}
    input: z $\in \mathbb{R} \sim \mathcal{N}(0,1); y \in \mathbb{R}$ & input: data sample $x \in \mathbb{R}^2$ \\\hline
    concat(z,y) $\in \mathbb{R}^2$ & Dense(100);ReLU \\\hline
    Dense(16);BN & Dense(100);ReLU \\\hline
    Dense(32);BN & Dense(100);ReLU \\\hline
    Dense(64);BN & Dense(100);ReLU \\\hline
    Dense(128);BN & Dense(100);ReLU \\\hline
    Dense(256);BN & Dense(1);Sigmoid \\\hline
    Dense(2) &  \\\specialrule{2pt}{0pt}{0pt}
    \end{tabular}
    \label{tab:untransformed_net}
\end{table}
\begin{table}[H]
    \centering
    \caption{Network architecture of the generator and discriminator for the examples detailed in section \ref{sec:gan-assisted} and \ref{sec:dim_and_extrap}. This is shown in the following format: layer type (output shape) ;nonlinearities where ``BN'' represents batch normalization. }
    \begin{tabular}{!{\VRule[2pt]}c!{\VRule}c!{\VRule[2pt]}}
    \specialrule{2pt}{0pt}{0pt}
    Generator & Discriminator\\\specialrule{2pt}{0pt}{0pt}
    input: z $\in \mathbb{R}^n \sim \mathcal{N}(0,1); y \in \mathbb{R}$ & input: data sample $x \in \mathbb{R}^m$ \\\hline
    concat(z,y) $\in \mathbb{R}^{n+1}$ & Dense(100);ReLU \\\hline
    Dense(200);BN;LeakyReLU & Dense(100);ReLU \\\hline
    Dense(200);BN;LeakyReLU & Dense(100);ReLU \\\hline
    Dense(200);BN;LeakyReLU & Dense(100);ReLU \\\hline
    Dense(200);BN;LeakyReLU & Dense(100);ReLU \\\hline
    Dense(200);BN;LeakyReLU & Dense(1);Sigmoid \\\hline
    Dense(m) &  \\\specialrule{2pt}{0pt}{0pt}
    \end{tabular}
    \label{tab:transformed_net}
\end{table}
\begin{table}[H]
    \centering
    \caption{Network architecture of the generator and discriminator for the example detailed in section \ref{sec:alanine}. This is shown in the following format: layer type (output shape) ; nonlinearities where ``BN'' represents batch normalization. }
    \begin{tabular}{!{\VRule[2pt]}c!{\VRule}c!{\VRule[2pt]}}
    \specialrule{2pt}{0pt}{0pt}
    Generator & Discriminator\\\specialrule{2pt}{0pt}{0pt}
    input: z $\in \mathbb{R}^n \sim \mathcal{N}(0,1); y \in \mathbb{R}$ & input: data sample $x \in \mathbb{R}^m$ \\\hline
    concat(z,y) $\in \mathbb{R}^{n+1}$ & Dense(100);ReLU \\\hline
    Dense(64);BN;LeakyReLU & Dense(200);ReLU \\\hline
    Dense(128);BN;LeakyReLU & Dense(200);ReLU \\\hline
    Dense(256);BN;LeakyReLU & Dense(200);ReLU \\\hline
    Dense(512);BN;LeakyReLU & Dense(200);ReLU \\\hline
    Dense(1024);BN;LeakyReLU & Dense(1);Sigmoid \\\hline
    Dense(m) &  \\\specialrule{2pt}{0pt}{0pt}
    \end{tabular}
    \label{tab:alanine_net}
\end{table}
In practice, GANs can suffer mode collapse during training. In the original CcGAN paper, the authors mention that mode collapse occurs when the cutoff distance specified in the loss function is too narrow \cite{ding2021ccgan}. We generally followed the guidelines from the CcGAN paper and did not experience overfitting or mode collapse during training. Early stopping of training upon convergence was also occasionally utilized to achieve optimal results.

\section{Diffusion Maps}
\label{sec:app-c}
Introduced by Coifman and Lafon \cite{COIFMANdmaps} and discussed in sections \ref{sec:US_on_dmap_CVS} and \ref{sec:gan-assisted} of this work, diffusion maps parametrize a data set of points \textbf{X} = $\{x_i\}^N_i$ sampled from a manifold where $\textbf{x}_i \in \mathbb{R}^m$, subsequently uncovering the geometry of the manifold and intrinsic geometry of the data. This is accomplished by first constructing an affinity matrix $\textbf{A} \in \mathbb{R}^{N \times N}$ by using a kernel function, often the Gaussian kernel:
\begin{equation*}
    A_{ij} = \exp\left\{-\frac{\| x_i - x_j \|^2}{2\epsilon}\right\},
\end{equation*}
where $||\cdot||$ denotes a norm (in our work, we use the Euclidean norm) and $\epsilon$ is a hyperparameter which regulates the rate of decay of the kernel. $\textbf{A}$ is then normalized in order to compute the intrinsic dimensionality of $\textbf{X}$ regardless of the sampling density by defining the diagonal matrix $\textbf{P}$ as follows:
\begin{equation*}
    P_{ii} = \sum_{j=1}^{N} A_{ij}
\end{equation*}
and renormalizing $\textbf{A}$ with
\begin{equation*}
    \Bar{\textbf{A}} = \textbf{P}^{-\alpha/2}\textbf{A}\textbf{P}^{-\alpha/2}
\end{equation*}
where $\alpha \in [0, 1]$ is our renormalization parameter (see~\cite{COIFMANdmaps}). $\Bar{\textbf{A}}$ is then further normalized to construct a row\--stochastic matrix $\textbf{W}$ with components,
\begin{equation*}
    W_{ij} = \frac{\Bar{A}_{ij}}{\sum_{j=1}^{N} \Bar{A}_{ij}}.
\end{equation*}
The eigendecompostion of \textbf{W} results in a set of eigenvalues and eigenvectors. The leading nonharmonic eigenvectors provide a (nonlinear) parametrization of \textbf{X} in diffusion map coordinates, once properly selected.\\
\indent In summary, diffusion maps are a data-driven technique that provide non-linear variables that parametrize a (possibly lower-dimensional) manifold of the inspected system. In our work, the first nontrivial eigenvector was one-to-one with the slow variable of the inspected dynamical systems; however, this is not always the case. If the first nontrivial eigenvector does not produce the desired CV, further inspection of other nonharmonic eigenvectors should be performed. Additionally, data-driven techniques such as Diffusion Maps in practice require analysis of a sufficiently large amount of data in order to identify diffusion map coordinates that parametrize a slow manifold of the system, and for all demonstrations of diffusion maps in this work, we performed diffusion maps on at least 5000 samples of data points from the dynamical system.

\section{Simulation and benchmarking details}
\label{sec:app-d}
 All simulations, calculations, and benchmarking were performed using an 8-core Intel Xeon E5-1660 v4 CPU. All systems of SDEs were integrated using the Euler--Maruyama method \cite{maruyama1955continuous} with time steps of dt = $1$ for the example in sections \ref{sec:US_on_dmap_CVS} and \ref{sec:cGAN_sampling}, dt = $5e-2$ for section \ref{sec:gan-assisted} (the equations of which are in Appendix \ref{sec:app-e}), and dt = $1e-3$ for section \ref{sec:dim_and_extrap}. All molecular dynamics simulations were performed using the OpenMM molecular dynamics package with the Amber14 force field in implicit solvent \cite{openmm, amber14, GBsolvent}. Stochastic (Langevin) dynamics were performed at 300 Kelvin with a time step of 1 fs and an inverse friction constant of 1 ps$^{-1}$. All US simulations were performed using OpenMM with the open-source PLUMED library \cite{plumed2019promoting, TRIBELLO_plumed2, plumed3}.

\section{It\=o's lemma application to the transformation of SDEs}
\label{sec:app-e}
Consider the SDE for $X(t,\omega) = (u(t,\omega),v(t,\omega))$,
\begin{equation*}
    \begin{split}
    &dx_1(t) = a_1dt + a_2dB_1(t),\\
    &dx_2(t) = a_3(1-x_2)dt + a_4dB_2(t),
    \end{split}
\end{equation*}
where $(B_1(t,\omega), B_2(t,\omega))$ is a standard two-dimensional Brownian motion.\par
Recall It\^o's lemma \cite{ito}, which says that if $f = f(X)$ is a deterministic function, then the stochastic process $Y(t,\omega) = (x(t,\omega),y(t,\omega)) = f(X(t,\omega))$ is governed by the SDE,
\begin{equation*}
    dY = (Df^T\mu+\frac{1}{2}\text{trace}(\sigma^TD^2f\sigma))dt+Df^T\sigma dB,
\end{equation*}
where $Df$ and $D^2f$ are the Jacobian and Hessian matrices of $f$, respectively. In our case,
\begin{equation*}
    f(x_1,x_2) = (x_2\cos{(x_1+x_2-1)}, x_2\sin{(x_1+x_2-1)}),
\end{equation*}
so we have the following SDEs post-transformation:
{\scriptsize
\begin{equation*}
    \begin{split}
    dy_1(t) = &- \frac{1}{2}\frac{((a_2^2+a_4^2+2a_3)y_1+2y_2(a_1+a_3))\sqrt{y_1^2+y_2^2}-2a_3y_1^2y_2-2a_3y_2^3+2a_4^2y_2-2a_3y_1}{y_1\sqrt{\frac{y_1^2 + y_2^2}{y_1^2}}}dt\\ 
    &- \frac{\sqrt{y_1^2+y_2^2} y_2a_2}{y_1\sqrt{\frac{y_1^2 + y_2^2}{y_1^2}}}dB_1 - \frac{1}{2}\frac{(2y_2a_4\sqrt{y_1^2+y_2^2}-2a_4y_1)}{y_1\sqrt{\frac{y_1^2 + y_2^2}{y_1^2}}}dB_2,\\
    dy_2(t) = &\frac{((a_1+a_3)y_1-\frac{1}{2}y_2(a_2^2+a_4^2+2a_3))\sqrt{y_1^2+y_2^2}-a_3y_1^3+(-a_3y_2^2+a_4^2)y_1+a_3y_2}{y_1\sqrt{\frac{y_1^2 + y_2^2}{y_1^2}}}dt\\ 
    &+\frac{a_2\sqrt{y_1^2+y_2^2}}{\sqrt{\frac{y_1^2 + y_2^2}{y_1^2}}}dB_1 + \frac{a_4\sqrt{y_1^2+y_2^2}y_1+y_2a_4}{\sqrt{y_1\frac{y_1^2 + y_2^2}{y_1^2}}}dB_2.\\
    \end{split}
\end{equation*}
}
The SDEs with the added US biases are
{\scriptsize
\begin{equation*}
    \begin{split}
    dy_1(t) = \biggl(- &\frac{1}{2}\frac{((a_2^2+a_4^2+2a_3)y_1+2y_2(a_1+a_3))\sqrt{y_1^2+y_2^2}-2a_3y_1^2y_2-2a_3y_2^3+2a_4^2y_2-2a_3y_1}{y_1\sqrt{\frac{y_1^2 + y_2^2}{y_1^2}}}\\&-k(\Phi_1((y_1(t),y_2(t))-\phi_0)\frac{\partial \Phi_1}{\partial y_1} \biggr) dt 
    - \frac{\sqrt{y_1^2+y_2^2} y_2a_2}{y_1\sqrt{\frac{y_1^2 + y_2^2}{y_1^2}}}dB_1 - \frac{1}{2}\frac{(2y_2a_4\sqrt{y_1^2+y_2^2}-2a_4y_1)}{y_1\sqrt{\frac{y_1^2 + y_2^2}{y_1^2}}}dB_2,\\
    dy_2(t) = \biggl(&(\frac{((a_1+a_3)y_1-\frac{1}{2}y_2(a_2^2+a_4^2+2a_3))\sqrt{y_1^2+y_2^2}-a_3y_1^3+(-a_3y_2^2+a_4^2)y_1+a_3y_2}{y_1\sqrt{\frac{y_1^2 + y_2^2}{y_1^2}}}\\&-k(\Phi_1((y_1(t),y_2(t))-\phi_0)\frac{\partial \Phi_1}{\partial y_2} \biggr)dt 
    +\frac{a_2\sqrt{y_1^2+y_2^2}}{\sqrt{\frac{y_1^2 + y_2^2}{y_1^2}}}dB_1 + \frac{a_4\sqrt{y_1^2+y_2^2}y_1+y_2a_4}{\sqrt{y_1\frac{y_1^2 + y_2^2}{y_1^2}}}dB_2.\\
    \end{split}
\end{equation*}
}

\bibliographystyle{unsrt}
\bibliography{bibliography}

\end{document}